\documentclass[10pt,twocolumn,letterpaper]{article}

\usepackage{iccv}
\usepackage{times}
\usepackage{epsfig}
\usepackage{graphicx}
\usepackage{amsmath}
\usepackage{amssymb}
\usepackage{booktabs}
\usepackage{array}
\usepackage{diagbox}
\usepackage{caption}
\usepackage{authblk}
\newcolumntype{C}[1]{>{\centering\arraybackslash}p{#1}}

\usepackage[breaklinks=true,bookmarks=false]{hyperref}

\iccvfinalcopy 

\def\iccvPaperID{} 

\ificcvfinal\fi

\begin{document}

\title{Are Adversarial Robustness and Common Perturbation Robustness Independent Attributes ?}

\author[1,2]{Alfred LAUGROS}
\author[1]{Alice CAPLIER}
\author[2]{Matthieu OSPICI}

\affil[1]{(Univ. Grenoble Alpes)}
\affil[2]{(Atos)}
\affil[ ]{\textit{\{alfred.laugros,alice.caplier\}@grenoble-inp.fr}}
\affil[ ]{\textit{matthieu.ospici@atos.net}}

\maketitle
\ificcvfinal\thispagestyle{empty}\fi

\begin{abstract}

Neural Networks have been shown to be sensitive to common perturbations such as blur, Gaussian noise, rotations, etc. They are also vulnerable to some artificial malicious corruptions called adversarial examples. The adversarial examples study has recently become very popular and it sometimes even reduces the term "adversarial robustness" to the term "robustness". Yet, we do not know to what extent the adversarial robustness is related to the global robustness. Similarly, we do not know if a robustness to various common perturbations such as translations or contrast losses for instance, could help with adversarial corruptions. We intend to study the links between the robustnesses of neural networks to both perturbations. With our experiments, we provide one of the first benchmark designed to estimate the robustness of neural networks to common perturbations. We show that increasing the robustness to carefully selected common perturbations, can make neural networks more robust to unseen common perturbations. We also prove that adversarial robustness and robustness to common perturbations  are independent. Our results make us believe that neural network robustness should be addressed in a broader sense.

\end{abstract}

\section{Introduction}

Deep Neural Networks have been shown to be very efficient in image processing tasks such as content classification \cite{imagenet}, face recognition \cite{face_rec_ref} or object detection \cite{rcnn}. Despite their good performances on academic datasets, artificial neural networks are vulnerable to common perturbations like blur, lightning variations or colorimetry changes \cite{face_rec_noise,noise_vul}. These perturbations are often encountered in industrial applications and can make some models useless.

Some techniques can be used to increase neural network robustness to such perturbations. Data augmentation approaches \cite{augmentation_doc} or fine tuning techniques \cite{fine_tune_noise} are broadly used to protect neural networks. Regularization techniques in general are useful to build models robust to traditional perturbations \cite{regular_noise}. Despite recent great advances, neural networks are not successful enough at dealing with corrupted images coming from real world applications \cite{human_vs_noise}.

Deep neural networks are also vulnerable to slightly modified samples called adversarial examples \cite{ref_adv}. They consist in an addition of malicious patterns that can completely disturb the behavior of a neural network.

A lot of defense strategies have been proposed to make neural networks more robust to adversarial examples. Distillation learning can be used to make neural networks more stable \cite{distillation_adversarial}. Introducing adversarial examples in the training procedure can decrease neural networks sensitivity to these attacks \cite{pgd_aug}. Additional modules such as autoencoder \cite{autoenc_adv}, or GAN \cite{defense_gan}, have been used to protect neural networks from adversarial corruptions. Regularization is also a standard procedure to make neural networks more robust to adversarial samples \cite{regular_adv}. However, none of these techniques succeeds in making a neural network perfectly invariant to adversarial examples.

In some recent studies, the expression "noise robustness" or the expression "adversarial robustness" are reduced to "robustness" only \cite{c&w,random_self_defense}. However, we do not know to what extent these fields are linked. Some recent works show that the salient points used by adversarially trained models to understand images are close to the ones used by humans \cite{odds_with}. Then, one could expect these models to be robust to the same kinds of perturbations as humans. We wonder if making neural networks more robust to adversarial perturbations could also make them more robust to common perturbations and vice versa. To better understand neural network global robustness, we want to study the hypothetical correlations between different kinds of robustnesses.

As part of our studies, we provide a thorough method to build a set of common perturbations in order to estimate the robustness of neural networks. We carry out experiments about making neural networks robust to unseen common perturbations. Finally, we study eventual correlations between the adversarial robustness and the robustness to common perturbations.

\section{Background and Related Works}

\subsection{Common perturbations}

A few academic datasets are used to compare state of the art networks: \cite{imagenet,celeba,coco}. Academic datasets are really useful to researchers, but they do not necessarily cover the various perturbations encountered in real application cases. In the CelebA dataset for instance, lightning conditions, face positions and colorimetry are constant \cite{celeba}. But in real application cases, various transformations can be introduced by sensor characteristics, lighting conditions or motions. Image processing or data transmission can also introduce unexpected distortions. We call these distortions common perturbations. We consider that common perturbations are transformations that are often encountered in industrial contexts but which are generally absent from academic datasets. This definition includes traditional additive noises such as Gaussian or salt-pepper noises. It also covers global changes in lightning conditions, contrast or colorimetry. Geometric transformations (translations, rotations...) are also included in this definition.

\subsection{Adversarial Examples} \label{adversarial_attacks}

Adversarial perturbations are small artificial corruptions introduced into clean samples so as to fool deep neural networks. A lot of attacks can potentially harm most of the neural networks: \cite{harnessing,c&w,adv_phy}. In our study, we choose to consider four of the most broadly used attacks: FGSM, PGD, LL-FGSM and LL-PGD. They are efficient, and can be easily computed to test the neural network robustness.

FGSM (Fast Gradient Sign Method) is one of the simplest method used to build adversarial examples \cite{harnessing}:

\begin{equation} \label{fgsm_expr}
x_{adv} = x + \epsilon*sign(\nabla_xL(x,l_{true}))
\end{equation}

With $x$ a sample to transform, $l_{true}$ the corresponding label and $L$ the cost function of the model. We note $\epsilon$ the amount of introduced adversarial perturbation.

PGD (Projected Gradient Descent) is an iterative version of FGSM \cite{adv_phy}:

\begin{equation} \label{pgd_expr}
x^{k+1} = x^{k} + \frac{\epsilon}{n}*sign(\nabla_xL(x^k,l_{true}))
\end{equation}

Starting with $x^0=x$, the upper expression is computed $n$ times to craft an adversarial example.

Instead of increasing the value of the loss function, it is possible to target a class in order to make a neural network associate the received sample with the targeted class. In particular, LL-FGSM (Least Likely FGSM) is a variation of FSGM that targets the class for which the targeted neural network has given the lowest score \cite{adv_phy}. Considering $l_{least}$, the label corresponding to the lowest score given by a neural network, an adversarial example is crafted by computing the following formula:

\begin{equation}
x_{adv} = x - \epsilon*sign(\nabla_xL(x,l_{least}))
\end{equation}

Similarily to LL-FGSM, LL-PGD is a variation of PGD that intends to make neural networks give a high score to the label $l_{least}$ \cite{adv_phy}. Starting with $x^0=x$, the adversarial example is built by computing several times the following expression:

\begin{equation}
x^{k+1} = x^{k} - \frac{\epsilon}{n}*sign(\nabla_xL(x^k,l_{least}))
\end{equation}

When the targeted model is known (we have access to its architecture and weights), we can directly use it to compute the gradient required for an adversarial attack. In this case, it is a white-box attack. These attacks are particularly harmful because they are built specifically to fool a precise model.

When the targeted model is unknown, the gradient used for the adversarial example crafting is computed with a different model. This is called a black-box attack. The model used for the gradient computing has a different architecture and different weights than the attacked model. Adversarial examples are transferable among neural networks \cite{ref_adv}. It means that attacks built on a network generally fool other networks, even if they are very different. Then, black-box attacks remain harmful on most models. Making a neural network robust to any black-box attack is a challenging issue.

\subsection{Data Augmentation}

Data augmentation is a technique used to make a neural network more robust to a kind of perturbation. It consists in introducing corrupted samples into the training set. In other words, a neural network that is trained with data augmentation, learns on clean samples but also on samples modified with a perturbation. At the end of the training, the augmented neural network has been made robust to the perturbation used during the training \cite{augmentation_doc}. When the perturbation used during the training is an adversarial attack, the data augmentation procedure is called adversarial training \cite{harnessing}.

\subsection{Robustness}

The definition of robustness is not a consensus. It can refer to different concepts depending on the context. In this paper, the notion of robustness is considered in relation with some perturbations. We explicitly indicate the perturbations towards which the robustness is considered. For instance we study the robustness to translations or the robustness to adversarial examples etc.

We note $A_{clean}$, the accuracy of the neural network $N$ on a test set. We consider some perturbations $\phi$ which are used to modify the samples of this test set. $A_{\phi}$ is the accuracy of the model on the test set modified with a $\phi$ perturbation. We measure the robustness of $N$ to a $\phi$ perturbation with the expression:

\begin{equation} \label{robust_expr}
R^{\phi}_{N} = \frac{A_{\phi}}{A_{clean}}
\end{equation}

We call it a robustness score and it measures the accuracy loss due to the $\phi$ perturbation. The more the robustness score of a model is close to one, the more it is robust to the considered perturbation. To measure the robustness score of a neural network to a set of perturbations S, we use:

\begin{equation}
R^{S}_{N} = \sum_{\phi \in S}{R^{\phi}_{N}}
\end{equation}


\subsection{Related Works}

\textbf{Benchmark to estimate the robustness of neural networks}.
\\
In \cite{noise_bench}, the ImageNet-C benchmark is used to measure the robustness of neural networks to common perturbations. Their benchmark is built on a set of common perturbations on which neural networks should be tested. Unfortunately, some kinds of perturbations in the set are over-represented and some others are not taken into account. In particular, ImageNet-C contains three kinds of noises and four kinds of blurs, but occlusions, translations and rotations are not present in it. Then, we decided to build another benchmark, more representative of common perturbations encountered in real application cases. The method used to build it is given in Section \ref{bench constr}.

The robustness to adversarial examples of a neural network, is usually measured by testing the performances of this network against several kinds of adversarial attacks \cite{ensemble_adversarial,adv_scale,defense_gan}.
\\
\\
\indent \textbf{Links between adversarial robustness and robustness to common perturbations}. \\
A few works study the links between the adversarial robustness and some specific noise robustnesses. For instance, relations between adversarial perturbations and random noises are established in \cite{random_and_adv, lp_noise_adv}. These relations prove that neural networks can be robust to random noises and remain vulnerable to adversarial attacks.

Links between small geometric transformations and adversarial examples have been established in \cite{local_geom_adv,geom_adv}. It is argued that the robustness to additive adversarial perturbations and the robustness to rotations and translations are orthogonal concepts.

A few links between some common perturbation robustnesses and some adversarial attack robustnesses are established in these works. However, they compare adversarial robustness with the robustness to a few specific common perturbations. In this study, we consider robustness to common perturbations in a broad sense: most of the common perturbations are included. We want to know if the global adversarial robustness could help neural networks to be globally more robust to common perturbations and conversely. We intend to enlarge the scope of the previous works to know more about the way the neural network robustnesses are related to each other.


\section{Construction of the Common Perturbation Benchmark} \label{bench constr}

\subsection{Experiment Set-up} \label{exp_set_up}

We choose the ImageNet dataset to carry out our experiments \cite{imagenet}. ImageNet is widely used, challenging and big enough for achieving adversarial trainings \cite{small_set_adv}. Adversarial training and data augmentation are computationally expensive, they increase the training times of neural networks. To speed up the trainings, we decided to use a subset of ImageNet. It is composed of 5 super-classes, each regrouping several ImagetNet classes. The chosen classes are: \textit{bird}, \textit{dog}, \textit{insect}, \textit{primate} and \textit{fish}. They correspond respectively to the ImageNet class ranges 80-100, 151-268, 300-319, 365-382 and 389-397. The choice of the classes was made by drawing inspiration from the experiments conducted in \cite{odds_with}. We insure the size equality of the classes by splitting the biggest classes to fit the smallest one. The resulting classes each contain ten thousand images. The obtained dataset is more suitable for achieving dozens of trainings in a reasonable amount of time, without loosing generality regarding the robustness study we want to conduct.

We use the ResNet-18 and ResNet-50 neural networks \cite{resnet} for our studies. The results shown in the Tables of the paper have been obtained with ResNet-18. Yet, the same experiments conducted with ResNet-50 lead us to the same conclusions than the ones found with ResNet-18. Those models are trained to classify the images extracted from our ImageNet sub-dataset. We use a stochastic gradient descent with a learning rate of 0.01. The learning rate is divided by 10 when the training accuracy reaches a plateau. We use a weight decay of 0.0001 and a batch size of 128. The loss used is a cross-entropy function. We call the \textit{standard} model, a ResNet-18 trained with these hyperparameters without using any data augmentation. It has an accuracy $A_{clean}$ of 0.83 on the ImageNet sub-dataset. We also train a VGG network \cite{vgg} and use it to get the gradient for the black-box adversarial attacks. For the trainings and the tests introducing adversarial examples, the amount of the corruption ($\epsilon$) of each example,  is randomly chosen. It varies from 0.01 to 0.1 for image pixel values that range from -1 to 1.

\subsection{Perturbation Selection Criteria} 

To conduct the study, we need a method to estimate the robustness to common perturbations of neural networks. A natural way to do this is to estimate the network robustness to diverse kinds of common perturbations. The quality of the estimation greatly depends on the relevance of the chosen perturbations. We build a set of perturbations based on three selection criteria: completeness, virulence and non-overlapping.

\textbf{Completeness}. A complete set of common perturbations should cover most of the perturbations commonly encountered in real world applications. To be considered robust to common perturbations, a neural network should be robust to as many common perturbations as possible. To build the most exhaustive list of perturbations, we gather ideas from several sources. We choose perturbations encountered in various industrial applications: video surveillance, production line, autonomous driving, etc... We also get inspired by some other works \cite{noise_bench, face_rec_noise}. At this step, we obtain the following set of perturbations: Gaussian noise, salt-pepper noise, speckle noise, defocus blur, motion blur, zoom blur, glass blur, rotations, translations, vertical flips, obstructions, brightness variations, contrast loss, colorimetry variations, interference distortions, quantizations and jpeg compression.

The Gaussian noise may appear because of sensors high temperature or poor illumination during acquisition. Salt-pepper noise is generally due to errors caused by a conversion from an analog signal to a digital signal. Speckle noise often corrupts images captured by radars or medical imaging systems. Defocus blur appears because of bad camera focusing. Motion blur is caused by camera motions or displacement of observed objects. Zooms of cameras can introduce zoom blur. Glass blur is often observed because of translucent obstacles. The orientation and the position of observed objects can change depending on the context. For instance some pieces in a production line can be displaced or be inside out. We model this with translations, rotations and vertical flip transformations. Brightness, contrast and colorimetry vary with lightning conditions and sensors characteristics. Electrical interferences may appear during image capturing and perturb images. We model these interferences with small periodic artifacts. Quantization causes rounding errors that modify images. Some artifacts can appear because of jpeg compressions.

\textbf{Virulence}. Some perturbations are very virulent and disturb significantly neural networks. Some other corruptions are harmless: they do not cause a significative drop in model performances. Then, being robust to harmless perturbations is not an interesting attribute. We test the robustness of the \textit{standard} model with all the perturbations selected in the previous paragraph. Most of these perturbations are virulent. For instance, the robustness score measured for Gaussian noise is 0.81. However, for quantizations or jpeg compression distortions, the measured robustness scores are above 0.97. We consider both corruptions not virulent. They are removed from the set of perturbations.

\textbf{Non-overlapping}. The robustnesses to two distinct perturbations can be correlated. Making a neural network more robust to a perturbation can also make it more robust to another perturbation and conversely. In this case, we consider that these robustnesses overlap. The presence of some overlapping perturbation robustnesses can unbalance the sum computed with the formula (\ref{robust_expr}). If a kind of robustness is over-represented, it distorts the robustness measure.

We conduct experiments to show that the robustnesses to traditional noises overlap. We train three identical Resnet-18, respectively augmented with Gaussian noise, salt-pepper noise and speckle noise. For each model, its robustness score towards each noise is computed: results are presented in the left part of Table \ref{tab:noise_overlapping}. Each line of the table except the standard one refers to an augmented model. Each column of the table refers to the noise used to compute the robustness scores of the line. For instance, the model augmented with a speckle noise, tested on samples corrupted with a salt-pepper noise, has a robustness score of 0.96. Every table of the paper is built this way: the lines refer to various models, the columns refer to the perturbations to which the robustnesses of the networks are evaluated.

In our noise study, we observe that if a model is robust to either the Gaussian noise, the salt-pepper noise or the speckle noise, it is also robust to the others. Then, these noise robustnesses overlap and so only the Gaussian noise is kept for the study. 

Similarly, we train four identical Resnet-18, respectively augmented with defocus blur, zoom blur, motion blur and glass blur. Taking into account the robustness scores reported in the right part of Table \ref{tab:noise_overlapping}, it appears that different kinds of blur robustnesses also overlap. Therefore, only the defocus blur is considered in further studies. We simply call it blur. 

Blur and noise perturbations are the only perturbations of our list for which a robustness overlapping is observed.

\begin{table*}[h]
\begin{center}
\centering
\begin{tabular}{ccccc}
\begin{tabular}{|p{20mm}|C{8mm}C{8mm}C{8mm}|}
\toprule
\diagbox[width=22mm]{Model}{Noise} &  gaus &  salt &  speck \\
\midrule
standard       &   0.81 &         0.79 &     0.87 \\
gaussian    &   0.98 &         0.98 &     0.98 \\
salt\_pepper &  0.99 &         0.99 &     0.97 \\
speckle     &   0.97 &         0.96 &     0.99 \\
\bottomrule
\end{tabular}
&
&
&
&
\begin{tabular}{|p{20mm}|C{8mm}C{8mm}C{8mm}C{8mm}|}
\toprule
\diagbox[width=24mm]{Model}{Noise} & defo &  zoom &  glass &  motion \\
\midrule
standard         &     0.61 &       0.80 &        0.74 &         0.74 \\
defocus\_blur &    0.98 &       0.94 &        0.98 &         0.94 \\
zoom\_blur    &    0.89 &       0.98 &        0.93 &         0.93 \\
glass\_blur   &   0.96 &       0.94 &        0.98 &         0.95 \\
motion\_blur  &  0.90 &       0.93 &        0.95 &         0.99 \\
\bottomrule
\end{tabular}
\end{tabular}
\end{center}
\captionsetup{justification=centering}
\caption{Left. Robustness scores towards noise perturbations | Right. Robustness scores towards blur perturbations.\\ Each line title of the table refers to the model used to compute the robustness score. The line "salt-pepper" for instance, corresponds to a model for which the training set has been augmented with salt-pepper corrupted images. Each column name refers to the perturbation used to compute the robustness scores of the column. We observe correlations between punctual noise robustnesses and correlations between blur robustnesses.}
\label{tab:noise_overlapping}
\end{table*}

\begin{figure*}[h]
  \centering
  \includegraphics[scale=0.75]{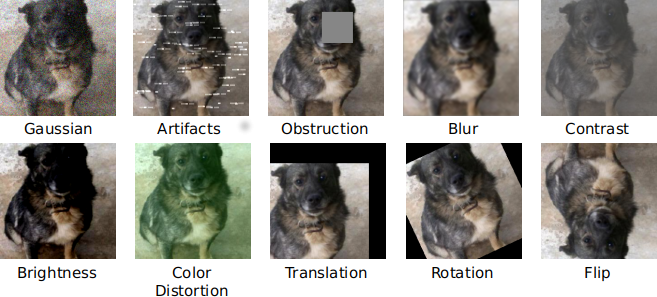}
  \captionsetup{justification=centering}
  \caption{Visualization of the selected common perturbations for the benchmark. These perturbations are fundamentally different by nature and affect distinct characteristics of images. They cannot be reduced to a smaller set without a significant loss of diversity. Each perturbation except \textit{flip} is provided with a continuous range of severity.}
  \label{fig:benchmark}
\end{figure*}

\textbf{Selected Perturbations}. Relying on those three criteria, we finally gather the set of common perturbations shown in Figure \ref{fig:benchmark}. From the upper left image of this Figure to the lower right one, the corresponding perturbations may be abbreviated with: \textit{gaus}, \textit{art}, \textit{obstr}, \textit{blur}, \textit{contr}, \textit{bright}, \textit{color}, \textit{trans}, \textit{rot} and \textit{flip}.

\subsection{Perturbation Intensities}

Each of the selected perturbations except \textit{flip} is associated with a range of intensity. For instance, square masks used for the occlusion perturbation can vary from 5 to 15 percent of the image size. We define a procedure to fix the upper and lower bounds of each perturbation intensity range. In order to fix the lower bound, the procedure starts with a very small perturbation. The intensity of the perturbation is progressively raised. During this increase, the behavior of the \textit{standard} model on the corrupted images is periodically tested. The lower bound of the perturbation range is reached when the accuracy of the neural network starts to decrease. We keep increasing the severity until the perturbation becomes visually disturbing for humans. At this point, the upper bound is fixed.

To our knowledge, this is the first set of common perturbations built on a such thorough method of selection. It is sufficiently complete to cover most of the widely spread perturbations. It can be used either in a data augmentation procedure or to estimate the robustness to common perturbations of a neural network. It is small enough to be used without increasing the computational cost of trainings and tests too much. We believe that using this set of perturbations, could help to build neural networks more stable when deployed in real environments.

\section{Common Perturbation Robustness and Adversarial Robustness}

\subsection{Robust Network Constructions}

To study the links between the adversarial robustness and the robustness to common perturbations, it is necessary to build neural networks robust to common perturbations on the one side, and neural networks robust to adversarial examples on the other side.

\textbf{A model robust to common perturbations}. We first train a ResNet-18 augmented with all the common perturbations of our set. We call it the \textit{fully augmented} model. We test its robustness to each perturbation of the set. The robustness scores obtained are compared with the ones obtained with the \textit{standard} ResNet-18: results can be found in Table \ref{tab:augmented_vs_standard}. As expected, the \textit{fully augmented} model is much more robust to any tested perturbation than the \textit{standard} one.

\begin{table*}[h]
\begin{center}
\centering
\begin{tabular}{|p{24mm}|C{8mm}C{8mm}C{8mm}C{8mm}C{8mm}C{8mm}C{8mm}C{8mm}C{8mm}C{8mm}C{8mm}|}
\toprule
\diagbox[width=22mm]{Model}{Noise} & gaus &  art &  obstr &  blur &  contr &  bright &  color &  trans &  rot &  flip &  mean \\
\midrule
standard   &  0.81 &       0.81 &         0.95 &  0.78 &      0.87 &        0.94 &             0.72 &         0.89 &      0.94 &           0.78 &  0.85 \\
fully augmented &  0.96 &       0.98 &         0.97 &  0.95 &      0.98 &        0.97 &             0.96 &         0.98 &      0.98 &           0.92 &  0.97 \\
\bottomrule
\end{tabular}
\end{center}
\captionsetup{justification=centering}
\caption{Efficiency of data augmentation on robustness. The robustness scores of the \textit{standard} and \textit{fully augmented} models are computed on the perturbations of the benchmark. Data augmentation makes the \textit{fully augmented} model much more robust to common perturbations than the \textit{standard} model.}
\label{tab:augmented_vs_standard}
\end{table*}

Data augmentation is supposed to make neural networks robust only to the distortions used in the augmentation process \cite{mixt_expert}. Then, there is no guarantee that the \textit{fully augmented} model is robust to common perturbations on which it has not been trained. We build a second experiment to guarantee that the \textit{fully augmented} network is more robust to any common perturbations than the \textit{standard} model. To achieve this, we train several ResNet-18, each augmented with all perturbations of our set but one. For instance, the \textit{no-gaussian} model is the model augmented with all perturbations but the Gaussian noise. In a general way, the \textit{no-$\phi$} model has been submitted to all perturbations of the set but \textit{$\phi$}. We compute the robustness of each \textit{no-$\phi$} model on samples corrupted with the \textit{$\phi$} perturbation. We compare every robustness score found this way with the ones of the \textit{standard} model: results are presented in Table \ref{tab:but_one}.
\begin{table*}[h]
\begin{center}
\centering
\begin{tabular}{|p{18mm}|C{8mm}C{8mm}C{8mm}C{8mm}C{8mm}C{8mm}C{8mm}C{8mm}C{8mm}C{8mm}|}
\toprule
\diagbox[width=22mm]{Model}{Noise} &  gaus &  art &  obstr &  blur &  contr &  bright &  color &  trans &  rot &  flip \\
\midrule
standard &   0.81 &       0.81 &         0.95 &  0.78 &      0.87 &        0.94 &             0.72 &         0.89 &      0.94 &           0.78 \\
no-$\phi$  &  0.83 &       0.84 &         0.96 &  0.82 &      0.94 &        0.97 &             0.76 &         0.92 &      0.95 &           0.78 \\
\bottomrule
\end{tabular}
\end{center}
\captionsetup{justification=centering}
\caption{Robustness to unseen perturbations. Each score of the second line refers to the robustness of a \textit{no-$\phi$} model against the \textit{$\phi$} perturbation. Various augmentations help the \textit{no-$\phi$} models to deal with the unseen \textit{$\phi$} perturbations.}
\label{tab:but_one}
\end{table*}

It appears that even if each \textit{no-$\phi$} model has never seen the \textit{$\phi$} perturbation, it is slightly more robust to it than the \textit{standard} model. It is true that the robustnesses to two very different distortions are usually not correlated: robustness to blur does not help with Gaussian noise. Yet, we observe that an increase in robustness to a group of common perturbations, can imply a better robustness to a very different perturbation. The \textit{no-gaussian} model is more robust to Gaussian noise than the \textit{standard} model. 

Therefore, with a sufficiently large and diverse set of common perturbations, it is possible to make a neural network more robust to an unseen common perturbation. As the \textit{fully augmented} model is trained on more perturbations than \textit{no-$\phi$} models, it has even more chances to be robust to any common perturbations. Consequently, in further experiments, the \textit{fully augmented} model is considered globally more robust to common perturbations than the \textit{standard} model.

\textbf{A model robust to adversarial examples}. To build a ResNet-18 robust to adversarial attacks, we use adversarial training. We call \textit{fgsm}, \textit{pgd}, \textit{ll\_fgsm} and \textit{ll\_pgd}, the models respectively augmented with FGSM, PGD, LL-FGSM and LL-PGD adversarial examples. To estimate their adversarial robustness, we compute their robustness scores on every attack introduced in section \ref{adversarial_attacks}. These scores are computed in black-box and white-box settings. Results are provided in Table \ref{tab:adversarial}. Compared to the \textit{standard} model, the adversarially trained models are more robust to every adversarial attack we test, in black-box and white-box settings. 

Besides, even if each of these models has been trained only with one kind of adversarial example, all of them are relatively robust to other kinds of adversarial examples. For instance the \textit{fgsm} model is more robust than the \textit{standard} model to LL-FGSM, PGD and LL-PGD adversarial examples. It means that adversarial example robustnesses are correlated: making a network robust to a specific adversarial attack, helps it to deal with other kinds of adversarial perturbations. 

\begin{table*}[h]
\begin{center}
\centering
\begin{tabular}{cccc}
\begin{tabular}{|p{24mm}|C{8mm}C{8mm}C{8mm}C{8mm}|}
\toprule
\diagbox[width=22mm]{Model}{Attack} &  fgsm &  fgsm\_ll &   pgd &  pgd\_ll \\
\midrule
standard &  0.68 &     0.70 &  0.73 &    0.95 \\
fgsm  &  0.96 &     0.97 &  0.97 &    99 \\
fgsm\_ll &   0.96 &     0.97 &  0.98 &    0.99 \\
pgd     &  0.98 &     0.98 &  0.98 &    1.00 \\
pgd\_ll  &   0.93 &     0.94 &  0.98 &    0.99 \\
fully augmented &  0.67 &     0.68 &  0.73 &    0.95 \\
\bottomrule
\end{tabular}
&
&
&
\begin{tabular}{|p{24mm}|C{8mm}C{8mm}C{8mm}C{8mm}|}
\toprule
\diagbox[width=22mm]{Model}{Attack} &  fgsm &  fgsm\_ll &   pgd &  pgd\_ll \\
\midrule
standard &  0.02 &     0.07 &  0.00 &    0.04 \\
fgsm  &  0.62 &     0.89 &  0.27 &    0.67 \\
fgsm\_ll &   0.42 &     0.75 &  0.36 &    0.79 \\
pgd     &   0.47 &     0.86 &  0.42 &    0.85 \\
pgd\_ll  & 0.41 &     0.79 &  0.38 &    0.82 \\
fully augmented &   0.02 &     0.08 &  0.01 &    0.05 \\
\bottomrule
\end{tabular}
\end{tabular}
\captionsetup{justification=centering}
\caption{Left. Robustness to black-box attacks | Right. Robustness to white-box attacks \\
The columns of the tables refer to the adversarial attacks used to perturb a model, either in a black-box (left) configuration or in a white-box configuration (right). The adversarially trained models are much more robust to adversarial examples than the \textit{standard} and \textit{fully augmented} models.
}
\label{tab:adversarial}
\end{center}
\end{table*}

\subsection{Links between Adversarial Robustness and Common Perturbation Robustness}

We can observe correlations between some common perturbation robustnesses. Likewise, increasing the robustness to an adversarial attack makes neural networks less sensitive to other adversarial perturbations. But are there correlations between adversarial robustnesses and robustness to common perturbations ?

We measure the robustness of the network augmented with common perturbations to adversarial examples and vice-versa. In Table \ref{tab:adversarial}, it appears that the \textit{fully augmented} model is not more robust than the \textit{standard} model to adversarial attacks: their robustness scores are almost equal. So, robustness to common perturbations does not protect from adversarial attacks. Similarly, as showed in Table \ref{tab:from_adv_to_noise}, the adversarially trained models are not more robust to the common perturbations than the \textit{standard} model. Increasing the robustness to adversarial examples does not increase the robustness to common perturbations. Therefore, adversarial robustness and robustness to common perturbations are independent attributes.

\begin{table*}[h]
\begin{center}
\centering
\begin{tabular}{|p{18mm}|C{8mm}C{8mm}C{8mm}C{8mm}C{8mm}C{8mm}C{8mm}C{8mm}C{8mm}C{8mm}C{8mm}|}
\toprule
\diagbox[width=22mm]{Model}{Noise} &  gaus &  art &  obstr &  blur &  contr &  bright &  color &  transl &  rot &  flip &  mean \\
\midrule
standard &   0.81 &       0.81 &         0.95 &  0.78 &      0.87 &        0.94 &             0.72 &         0.89 &     0.94 &   0.78 &      0.85 \\
fgsm &   0.91 &       0.89 &         0.93 &  0.91 &      0.78 &        0.87 &             0.69 &         0.82 &     0.91 &   0.78 &      0.85 \\
fgsm\_ll &   0.96 &       0.87 &         0.95 &  0.91 &      0.73 &        0.88 &             0.67 &         0.83 &      0.92 &    0.78 &  0.85 \\
pgd     &  0.95 &       0.88 &         0.94 &  0.94 &      0.75 &        0.87 &             0.69 &         0.81 &      0.91 &           0.79 &  0.85 \\
pgd\_ll  &  0.94 &       0.85 &         0.95 &  0.93 &      0.75 &        0.86 &             0.61 &         0.83 &      0.92 &           0.79 &  0.84 \\
\bottomrule
\end{tabular}
\end{center}
\captionsetup{justification=centering}
\caption{Robustnesses of adversarially trained models to common perturbations. The adversarially trained models are not more robust to common perturbations than the \textit{standard model}.
}
\label{tab:from_adv_to_noise}
\end{table*}

\section{Discussions}

We showed that relations exist between robustnesses to common perturbations. It also exists correlations between adversarial examples robustnesses. However adversarial robustness and common perturbation robustness are not correlated.

This discrepancy could be explained by the significant difference of nature between adversarial perturbations and common perturbations. An adversarial perturbation is an attack. It is made to disturb neural networks and it depends on the sample it corrupts: see formula (\ref{fgsm_expr}). Adversarial examples are based on an addition procedure and they are made with very small perturbations. On the other hand, common perturbations do not adapt to the neural networks they affect. They are not necessarily additive and can be very severe. Therefore, the way common perturbations and adversarial perturbations affect neural networks might be drastically different.

Recent works show the difference of nature between two kinds of features of images called \textit{robust} and \textit{non-robust features} \cite{adv_are_features}. The \textit{none-robust features} are the features exploited by adversarial examples to disturb the neural networks they attack. The \textit{robust features} are the features that are not modified by adversarial perturbations. In \cite{adv_are_features}, it is shown that adversarially trained models rely on \textit{robust features} while the models not augmented with adversarial perturbations rely on \textit{non-robust features} in addition to the \textit{robust features}. We think the differences of features used by these networks could explain the independence between adversarial robustness and robustness to common perturbations.

Be that as it may, robustness cannot be reduced to the sole adversarial robustness or the sole robustness to common perturbations. Those robustnesses are too poorly related. This independence is restrictive for industrial applications. Both robustnesses have to be addressed and addressed independently. It means that it would be necessary to introduce more regularization techniques or diversify the samples used in the augmentation procedures. This solution is time-consuming and computationally expensive.

From a theoretical point of view, we expect robust neural networks to extract relevant features. These features should be stable and should not change with common perturbations or adversarial attacks. Then, methods to make neural networks more robust should not depend on the nature of the corruption. They should naturally cover most common perturbations and adversarial attacks at the same time. Then, we believe that neural network robustness should be addressed in a broader sense: covering adversarial examples, common perturbations or even unexpected kinds of distortions simultaneously.

A possible approach to address robustness in a broad sense is to use alternative adversarial example definitions. Some new formulations have been proposed to enlarge the scope of adversarial perturbations \cite{geom_adv}, \cite{adv_render}. These definitions include small rotations, translations or lightning variations into the adversarial attack scope. Using more general adversarial examples in trainings could be a way to erase the discrepancy between the adversarial robustness and the robustness to common perturbations. If we can find a wide enough definition of adversarial attacks, both robustnesses could be addressed simultaneously.

Generative Adversarial Networks have been used to increase the adversarial robustness of some neural networks \cite{gan_trainer}. A generator is used to submit disturbing samples to another trained neural network. The generator is supposed to target the weaknesses of the other model. It automatically finds relevant attacks that help the other neural network to increase its robustness. The advantage of GAN is that the allowed range of perturbations is almost unlimited.  Perturbations introduced by the generator are not made by hand: they are not restricted to a few common perturbations or to some adversarial attacks. Generated attacks can contain a complex mix of common perturbations and adversarial patterns. GAN should be a good solution to introduce automatically a lot of kinds of perturbations in order to address robustness in a broader sense.

\section{Conclusion}

We carried out original experiments to better understand the links between the neural network robustnesses to different kinds of perturbations. We propose a new benchmark to estimate robustness to common perturbations. We showed that using data augmentation with a carefully chosen set of common perturbations, can increase the robustness of a model to an unknown common perturbation. We also demonstrated that adversarial robustness and robustness to common perturbations are independent attributes. We believe that the key to address neural network robustness in a broad sense, is to enlarge the scope of the perturbations used in trainings and tests, by considering corruptions that are in between common perturbations and adversarial attacks.


{\small
\bibliographystyle{ieee_fullname}
\bibliography{egbib}
}

\end{document}